\def\year{2022}\relax
\documentclass[letterpaper]{article} % DO NOT CHANGE THIS
\usepackage{aaai22}  % DO NOT CHANGE THIS
\usepackage{times}  % DO NOT CHANGE THIS
\usepackage{helvet}  % DO NOT CHANGE THIS
\usepackage{courier}  % DO NOT CHANGE THIS
\usepackage[hyphens]{url}  % DO NOT CHANGE THIS
\usepackage{graphicx} % DO NOT CHANGE THIS
\usepackage{natbib}  % DO NOT CHANGE THIS AND DO NOT ADD ANY OPTIONS TO IT
\usepackage{caption} % DO NOT CHANGE THIS AND DO NOT ADD ANY OPTIONS TO IT
\DeclareCaptionStyle{ruled}{labelfont=normalfont,labelsep=colon,strut=off} % DO NOT CHANGE THIS
\usepackage{booktabs}
\usepackage{makecell}
\usepackage[switch]{lineno}

\usepackage{multicol}
\usepackage{multirow}
\usepackage[lined,boxed,commentsnumbered,ruled]{algorithm2e}

\usepackage{amssymb}
\usepackage{amsmath}
\usepackage{color}

\let\oldequation\equation
\let\oldendequation\endequation

\title{Exploiting Fine-grained Face Forgery Clues via Progressive Enhancement Learning}

\author{}  %154 414 573  1009 1011

\author{
Qiqi Gu\textsuperscript{\rm 1 2}\footnote{Equal contribution.} \quad 
Shen Chen\textsuperscript{\rm 2}\footnotemark[\value{footnote}] \quad
Taiping Yao\textsuperscript{\rm 2}\footnotemark[\value{footnote}] \quad 
Yang Chen\textsuperscript{\rm 2} \quad
Shouhong Ding\textsuperscript{\rm 2}\footnote{Corresponding authors.} \quad 
Ran Yi\textsuperscript{\rm 1 3}\footnotemark[\value{footnote}]
}

\affiliations {
\textsuperscript{\rm 1}Shanghai Jiao Tong University, 
\textsuperscript{\rm 2}Youtu Lab, Tencent,
\textsuperscript{\rm 3}MoE Key Lab of Artificial Intelligence, SJTU\\
miemie@sjtu.edu.cn, \{kobeschen, taipingyao, wizyangchen, ericshding\}@tencent.com,  ranyi@sjtu.edu.cn 
}

\begin{document}
%\linenumbers
\maketitle

\begin{abstract}
With the rapid development of facial forgery techniques, forgery detection has attracted more and more attention due to security concerns. Existing approaches attempt to use frequency information to mine subtle artifacts under high-quality forged faces. However, the exploitation of frequency information is coarse-grained, and more importantly, their vanilla learning process struggles to extract fine-grained forgery traces. To address this issue, we propose a progressive enhancement learning  framework to exploit both the RGB and fine-grained frequency clues. Specifically, we perform a fine-grained decomposition of RGB images to completely decouple the real and fake traces in the frequency space. Subsequently, we propose a progressive enhancement learning framework based on a two-branch network, combined with self-enhancement and mutual-enhancement modules. The self-enhancement module captures the traces in different input spaces based on spatial noise enhancement and channel attention. The Mutual-enhancement module concurrently enhances RGB and frequency features by communicating in the shared spatial dimension. The progressive enhancement process facilitates the learning of discriminative features with fine-grained face forgery clues. Extensive experiments on several datasets show that our method outperforms the state-of-the-art face forgery detection methods.
\end{abstract}

\section{Introduction}

Over the past few years, face forgery technology has made significant progress. With public accessible tools such as deepfakes~\cite{deepfakes} or face2face~\cite{face2face}, people can easily manipulate the expression, attribution or identity of faces in images. Various powerful algorithms are utilized to generate realistic forged faces, which are hardly discerned by human eyes. These realistic forged faces have been used in pornography or political rumors, which are harmful to the community and evoke social panic. Under this background, the face forgery detection task was born and has attracted more and more attention recently.

Many significant works~\cite{xception, face_xray, df_mask, kobe, f3net, cvpr2021multi, guzhihao} have made great contributions to this task. 
Prior works use hand-crafted features (\emph{e.g.}, eye blinking~\cite{liy2018exposingaicreated}, head inconsistency~\cite{yang2019exposing}) or semantic features extracted by universal CNN \cite{mesonet, xception} to conduct binary classification.
Recently, two typical manners for mining subtle task-specific artifacts have been explored.
On the one hand, some works~\cite{df_mask, face_xray, kobe, pcl, wangxinyao} utilize auxiliary supervision such as blending boundary~\cite{face_xray} or forged mask~\cite{df_mask, kobe}. However, the mask ground-truth they rely on is difficult to access in most cases.
On the other hand, some attempt to employ the frequency-aware features to assist classification, using DCT~\cite{f3net}, local frequency~\cite{kobe} or SRM~\cite{cvpr2021highfreq}.
Although these works have made considerable performance, the frequency features extracted by their strategies are relatively coarse, which burdens the network to assemble discriminative feature patterns. Moreover, all these frequency-aware approaches either directly concatenate RGB and frequency features at the end of the network or fuse them only at a shallow layer. This makes it difficult for them to fully utilize the decomposed frequency information and limits the possibility of discovering fine-grained face forgery cues.

To address the above issues, we propose a novel Progressive Enhancement Learning (PEL) framework aiming at exploiting face forgery clues. Specifically, we employ the patch-wise Discrete Fourier Transform~\cite{learningFreq} with the sliding window to decompose the RGB input into fine-grained frequency components. Frequency domain information is assisted and complementary to the RGB features. And different from conventional frequency components, fine-grained frequency components can produce more combinations of features to discover potential artifacts.

To fully exploit RGB image and fine-grained frequency components, we design a two-stream network structure with two novel enhancement modules that progressively enhance the forgery clues. One is the self-enhancement module. It captures the traces in different input spaces separately based on spatial noise enhancement and channel attention. The other is the mutual-enhancement module, which achieves concurrent enhancement of both the RGB and frequency branches through feature communication in the shared spatial dimension. We insert these two enhancement modules after each convolutional block of the network. The earlier enhancements facilitate feature extractions in the later layers, and the forgery cues hidden in the fake faces are sufficiently excavated and magnified.

Extensive experiments are conducted to demonstrate the effectiveness of our method. With the progressive representation enhancement, better results are obtained on high-quality forged images.
Our method exceeds the other comparison methods and achieves state-of-the-art performance on the common benchmark FaceForensics++ dataset and the newly published WildDeepfake dataset. The cross-dataset evaluations on three additional challenging datasets prove the generalization ability, while the perturbed evaluations prove the robustness of our method.

Our contributions can be summarized as follows:
\begin{itemize}
    \item We propose a progressive enhancement learning network aiming at exploiting the combined fine-grained frequency components and RGB image.
    \item We first perform fine-grained decomposition in the frequency domain, and then propose two novel enhancement modules, \textit{i.e.}, self-enhancement and mutual-enhancement modules, which progressively enhance the feature to capture the subtle forgery traces.
    \item Extensive experiments and visualizations reveal the significance of feature enhancement in face forgery detection, and demonstrate the effectiveness of our proposed method against the state-of-the-art competitors. 
\end{itemize}

\section{Related work}
Due to the rapid development of face forgery methods, face forgery detection has recently attracted more attention in computer vision communities. 
Great efforts have been made to explore the authenticity of human faces.

Early works use hand-crafted features to seek for the artifacts in forged faces. Fridrich \textit{et al.}~\cite{steg_feature} extract steganalysis features and train a linear Support Vector Machine (SVM) classifier to distinguish authenticity. After that, Cozzolino \textit{et al.}~\cite{recasting} cast the features mentioned above to a convolution neural network. Matern \textit{et al.}~\cite{vis_arti} analyze the facial regions which are likely to generate artifacts and design specific descriptors to capture them. Yang \textit{et al.}~\cite{yang2019exposing} estimate the inconsistency of head poses between adjacent faces. With the rapid progress of deep learning, some CNN-based works~\cite{bayar2016deep, mesonet, xception} emerge to extract high-level semantic information for classification. However, their vanilla structures are unsuitable for the fast-growing and realistic forgery faces.

To further mine the heuristic forgery cues, two typical manners are widely used. One is resorting to auxiliary supervisions. For example, Nguyen \textit{et al.}~\cite{multi_task_df} incorporate the classification and segmentation task into one framework simultaneously.
Dang \textit{et al.}~\cite{df_mask} further highlight the manipulated area via attention mechanism. Face X-ray \cite{face_xray} focuses on the blending boundary induced by the image fusion process.
Besides, some works~\cite{kobe,pcl} resort to the similarity between local regions and achieve superior performance on benchmark datasets.
However, the manipulated mask above methods relied on training is not applicable to some wild datasets (e.g., Wild Deepfake dataset~\cite{wild_deepfake}), limiting practical application.

The others notice the artifacts hidden in the frequency domain that are difficult to observe directly.
Singhal \textit{et al.}~\cite{freq_cnn_image_manip} extracts frequency features through the residuals of image median filtering for detecting manipulated faces.
Frank \textit{et al.}~\cite{freq_gan} reveal the peculiar frequency spectrum pattern of GAN-generated deepfake images.
$F^3$-Net~\cite{f3net} employ the frequency information in two ways: image components which are transformed from a certain frequency band, and local frequency statistics calculated by frequency spectrum of patches.  %band of the frequency spectrum
Chen \textit{et al.}~\cite{kobe} use high-frequency components of the image to assist in calculating patch-similarities.
Luo \textit{et al.}~\cite{cvpr2021highfreq} use SRM filter to guide RGB features.
FDFL \cite{CVPR2021FDFL} utilize frequency information by DCT and adaptive frequency information mining block.
Although these methods have achieved remarkable performances, the exploitation of frequency information is coarse-grained. More importantly, their vanilla learning process is insufficient to make full use of the subtle artifacts in the frequency domain.
Different from existing works, we propose a novel progressive enhancement learning network that exploits the fine-grained frequency components.

\begin{figure*}[t]
\centering
\includegraphics[scale=0.64]{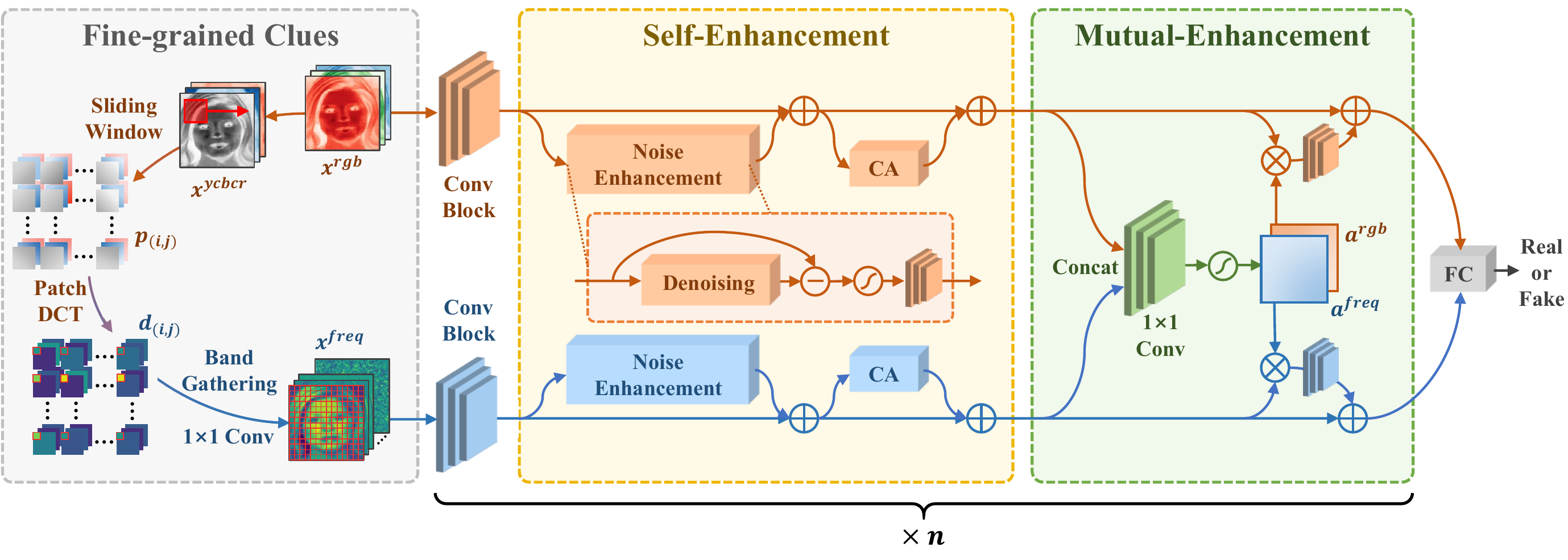}
\caption{The proposed PEL framework. The RGB input is decomposed into frequency components to mine fine-grained clues hidden in frequency space. In the self-enhancement module, the features of each stream are enhanced separately using a noise enhancement block and a channel attention block. In the mutual enhancement, both features are enhanced in the shared spatial dimension. The two enhancement modules are selectively inserted after each convolutional block of the network.}
\label{fig_net}
\end{figure*}

\section{Method}

Fig.~\ref{fig_net} illustrates the proposed PEL framework. We transform the RGB input into the fine-grained frequency component and fed them together into a two-stream network with EfficientNet as the backbone.
Each convolutional block of the backbone is followed by a self-enhancement module and a mutual-enhancement module to progressively enhance the fine-grained clues in an intra- and inter-stream manner.

\subsection{Fine-grained Clues}
\label{sec_toFreq}
As shown in the leftmost of Fig.~\ref{fig_net}, we apply domain transformation that decomposes the input RGB image into frequency components, revealing fine-grained forgery cues hidden in the frequency space.

Without loss of generality, let $x^{rgb} \in \mathbb{R}^{3 \times H \times W}$ denotes the RGB input, where $H$ and $W$ are the height and width of image. First, the RGB input $x^{rgb}$ is transformed to YCbCr space (denoted by $x^{ycbcr} \in \mathbb{R}^{3 \times H \times W}$) that coincides with the widely-used JPEG compression in the forged video.
Then, we slice $x^{ycbcr}$ via a sliding window to obtain a set of $8 \times 8$ patches. $p_{(i,j)}^c \in \mathbb{R}^{8 \times 8}$ denotes the patch sliced by the ($i$ th,$j$ th) sliding window of a certain color channel.
Each patch $p_{(i,j)}^c$ is processed by DCT into $8 \times 8$ frequency spectrum $d_{(i,j)}^c \in \mathbb{R}^{8 \times 8}$, where each value corresponds to the intensity of a certain frequency band. We flatten the frequency spectrum and group all components of the same frequency into one channel to form a new input, following the patch location in the original input:
\begin{equation}
    x^{flat}[:][i][j] = flatten(d_{(i,j)}),
\end{equation}
where $x^{flat} \in \mathbb{R}^{192 \times H_2 \times W_2}$ is the newly formed input, $H_2$ and $W_2$ are the vertical and horizontal step numbers of sliding window, and $d_{(i,j)} \in \mathbb{R}^{3 \times 8 \times 8}$ denotes the concatenation of all the $d_{(i,j)}^c$.
In this way, the original color input is decomposed and recombined to fine-grained frequency-aware data while maintaining the spatial relationship to match the shift-invariance of CNN. Finally, a $1 \times 1$ convolution layer is applied to the frequency data to adaptively extract the most informative frequency bands, forming the fine-grained frequency input $x^{freq} \in \mathbb{R}^{64 \times H \times W}$.

\subsection{Self-Enhancement Module}
\label{sec_self}

The self-enhancement module consists of two mountable sub-modules: a noise enhancement block and a channel attention block, which enhance the features of each stream separately. The visual artifacts in newly forged faces have significantly been eliminated~\cite{Li2019FaceShifterTH}, pushing subtle noise to a more critical role in the face forgery detection task. To this end, we design a noise enhancement block to excavate the artifact hidden in the feature noise, especially that of the fine-grained frequency input.

Given the RGB input $x^{rgb}$ and the frequency input $x^{freq}$, let $f_{in}^{rgb} \in \mathbb{R}^{c \times h \times w}$ and $f_{in}^{freq} \in \mathbb{R}^{c \times h \times w}$ denote the feature maps extracted by intermediate convolutional block of RGB stream and frequency stream, where $c$ is the number of channels, $h$ and $w$ denote the height and width of the feature maps, respectively. Since the RGB and frequency streams have the same network structure except for the first convolutional layer, we use the superscript $s \in \{rgb, freq\}$ to indicate different stream. 

For noise enhancement, we first extract the noise of input feature maps $f_{in}$ as follows:
\begin{equation}
    f_{noi}^{s} = f_{in}^{s} - \mathcal{M}(f_{in}^{s}),
\end{equation}
where $\mathcal{M}$ is the median filter with a kernel size of $3 \times 3$ that can filter noise such as salt to denoise at the feature level.
By subtracting the input features and denoising features spatially, $f_{noi}$ manifests the noise information of the input feature.

Then we apply a \emph{Sigmoid} function (denoted by $\sigma$) to amplify the subtle noises and suppress the exaggerated value of noise to prevent polarization. After that, a depth-wise $1 \times 1$ convolution layer is applied to the feature noise to adjust the amplitude of noise enhancement.
Finally, the feature noise is added to the input feature $f_{in}$ to obtain the noise-enhanced feature $f_{ne}$, which is used as input of the subsequent block. The above process can be summarized as follows:
\begin{equation}
    f_{ne}^{s} = f_{in}^{s} + Conv(\sigma(f_{noi}^{s})).
\end{equation}

Since the noise enhancement block considers the low-level noise of each channel separately, we further utilize a channel attention block to facilitate inter-channel interactions within each stream. 
Specifically, the channel attention block enhances features along the depth dimension, which combines the information from all channels and determines the significance of each channel, and enhances them accordingly.
The output feature $f_{out}$ is obtained as:
\begin{equation}
    f_{out}^{s} = f_{ne}^{s} + f_{ne}^{s} \otimes \sigma(MLP(GAP(f_{ne}^{s}) + GMP(f_{ne}^{s}))),
\end{equation}
where $MLP$ is a multi-layer perceptron, $GAP$ denotes the global average pooling, and $GMP$ is the global max pooling. 

In each stream of the network, the noise enhancement module is inserted after the first two convolutional blocks. This is because high-level features capture semantic information within a large reception field and are not suitable to extract the noise with local concepts. And the channel attention block is empirically mounted after the second to the penult convolutional blocks.
With the self-enhancement module, features are enhanced within the stream without changing their size, providing more helpful features for the following mutual-enhancement module.

\subsection{Mutual-Enhancement Module}
\label{sec_mutual}
As the features of two streams are originated from different input spaces, they provide distinct cues that can complement each other.
Thus, we design a mutual-enhancement module to comprehensively integrate the dual-stream knowledge along spatial dimension to enhance the features.

Concretely, the dual feature maps are concatenated 
and go through a point-wise convolution and a \emph{Sigmoid} function to get two spatial attention maps for two streams, respectively:
\begin{equation}
    A = \sigma(Conv(Cat(f_{in}^{rgb}, f_{in}^{freq}))),
\end{equation}
where $A \in \mathbb{R}^{2 \times h \times w}$ is the spatial attention maps, $Conv$ denote the point-wise convolution layer, and $Cat$ concatenate the features along with the depth. The input feature of each stream is multiplied with its attention map (\emph{i.e.} $a^{rgb}$ and $a^{freq}$ which are the two channels of $A$), and then go through a depth-wise $1 \times 1$ convolution layer to adjust the enhancing intensity. Finally, we add the enhanced feature with the input feature as follows (the feature size remains the same):
\begin{equation}
    f_{out}^{s} = f_{in}^{s} + Conv(f_{in}^{s} \otimes a^{s}).
\end{equation}

The mutual-enhancement module is inserted after the second to the penult convolutional blocks and placed behind the self-enhancement module. In this way, we can highlight forged regions that benefit from the other stream while maintaining stream-specific fine-grained forgery clues, which result in more discriminative features. 

\subsection{Loss Function}
In the preceding blocks, the two features enhanced by the mutual-enhancement module are fed into the next convolution blocks. In the last block, the outputs of two streams are concatenated for classification. We adopt the widely-used Binary Cross-Entropy loss as the objective function:

\begin{equation}
    \mathcal{L} = -\frac{1}{n}\sum(y_{gt} \times log(y_{pred}) + (1 - y_{gt}) \times (log(1 - y_{pred})),
\end{equation}
where $y_{gt}$ is the ground truth label, and $y_{pred}$ denotes the prediction of the network. The parameters of the network are updated via the back-propagation algorithm.

\section{Experiment}

% Table generated by Excel2LaTeX from sheet 'paper'
\begin{table*}[t]
  \centering

    \begin{tabular}{l|cccccccc}
    \hline
    \multirow{2}{*}{\textbf{Methods\quad \quad \quad \quad \quad \quad \quad \quad \quad \quad \quad \quad }} & \multicolumn{2}{c}{\textbf{\quad \quad FF++ (HQ)\quad \quad \quad}} &       & \multicolumn{2}{c}{\textbf{\quad \quad FF++ (LQ})\quad \quad \quad} &       & \multicolumn{2}{c}{\textbf{\quad WildDeepfake} \quad \quad}\\
\cline{2-3}\cline{5-6}\cline{8-9}          & \textbf{Acc} & \textbf{AUC} &       & \textbf{Acc} & \textbf{AUC} &       & \textbf{Acc} & \textbf{AUC}\\
    \hline
    C-Conv \cite{bayar2016deep} & 82.97\% & -     &      & 66.84\% & -  &      & - & -\\
    %\hline  
    CP-CNN  \cite{CP-CNN}& 79.08\% & -    &       & 61.18\% & -  &      & - & -\\  
    %\hline 
    MesoNet \cite{mesonet}& 83.10\% & -     &      & 70.47\% & -  &      & - & -\\
    %\hline
    Xception \cite{xception} & 95.73\% & -     &      & 86.86\% & - &      &  79.99\%$^\dagger$  & 0.8886$^\dagger$ \\
    %\hline
    Face X-ray \cite{face_xray} & -     & 0.8735 &      & -     & 0.6160  &      & - & -\\
    %\hline
    Two-branch RN \cite{two_branch_rn} & 96.43\% & 0.9870 &      & 86.34\% & 0.8659  &      & - & -\\
    %\hline
    Add-Net \cite{wild_deepfake} &96.78\%  &0.9774  &      & 87.50\% & 0.9101 &      &  76.31\%$^\dagger$  & 0.8328$^\dagger$ \\
    %\hline
    F$^3$-Net \cite{f3net} & 97.52\% & 0.9810 &      & 90.43\% & 0.9330 &      & 80.66\%$^\dagger$ & 0.8753$^\dagger$ \\
    %\hline
    FDFL \cite{CVPR2021FDFL} & 96.69\% & 0.9930 &      & 89.00\% & 0.9240 &      & - & - \\
    %\hline
    Multi-Att \cite{cvpr2021multi} & 97.60\% & 0.9929 &      & 88.69\% & 0.9040 &      & 81.99\%$^\dagger$ & 0.9057$^\dagger$ \\
    %\hline
    Ours  & \textbf{97.63\%} & \textbf{0.9932} &      & \textbf{90.52\%} & \textbf{0.9428} &      & \textbf{84.14\%} & \textbf{0.9162} \\
    \hline
    \end{tabular}%
    \caption{Quantitative results on FaceForensics++ dataset with different quality settings, and WildDeepfake dataset.}
  \label{tab:ffpp}%
\end{table*}%

In this section, we first introduce the overall experimental setup, then present extensive experimental results to demonstrate the effectiveness and robustness of our approach, and finally visualize the enhanced fine-grained face forgery clues and class activation map.

\subsection{Experimental Setup}

\noindent\textbf{Datasets.} We adopt five widely-used public datasets in our experiments, \emph{i.e.}, FaceForensics++ \cite{ffpp}, WildDeepfake~\cite{wild_deepfake}, Celeb-DF~\cite{celeb_df}, DeepfakeDetection~\cite{dfd}, DeepfakeDetectionChallenge \cite{dfdc},
in which the former two are used for both training and evaluation, while the latter three for cross-dataset evaluation only.
\begin{itemize}
    \item \textbf{FaceForensics++} (FF++) is a large-scale benchmark dataset containing 1000 original videos from youtube and corresponding fake videos which are generated by four typical manipulation methods: \emph{i.e.} Deepfakes (DF)~\cite{deepfakes}, Face2Face (F2F)~\cite{face2face}, FaceSwap (FS)~\cite{faceswap} and NeuralTextures (NT)~\cite{NT}. Different levels of compressions are used on raw videos, and two counterparts are generated, \emph{i.e.} high quality (HQ) and low quality (LQ). We follow the official splits by using 720 videos for training, 140 videos for validation, and 140 videos for testing.
    \item \textbf{WildDeepfake} is a dataset with 7,314 face sequences entirely collected from the internet, in which 6,508 sequences are for training and 806 are for testing. The contents in this dataset are diverse in many aspects, \emph{e.g.} activities, scenes, and manipulation methods, which makes this dataset more challenging and closer to the real-world face manipulation scenario.
    \item \textbf{Celeb-DF} is composed of 590 real videos based on 59 celebrities from youtube, and the corresponding 5639 fake videos tampered with the improved DeepFake algorithm~\cite{celeb_df}.
    \item \textbf{DeepfakeDetection} (DFD) is a large dataset that contains over 3000 manipulated videos in various scenes using publicly available deepfake generation methods.
    \item \textbf{DeepfakeDetectionChallenge} (DFDC) is a prominent face swap video dataset, with over 100,000 total clips sourced from 3,426 paid actors, produced with several Deepfake, GAN-based, and non-learned methods.
\end{itemize}

\newcommand{\tabincell}[2]{\begin{tabular}{@{}#1@{}}#2\end{tabular}}

\begin{table*}[htbp]
  \centering
    \begin{tabular}{cl|ccccccccccc}
    \hline
    \multirow{3}{*}{\tabincell{c}{\textbf{Training}\\\textbf{dataset}} } & \multirow{3}{*}{\textbf{Methods}} & \multicolumn{11}{c}{\textbf{Testing dataset}} \\
\cline{3-13}          &       & \multicolumn{2}{c}{\textbf{WildDeepfake}} &       & \multicolumn{2}{c}{\textbf{Celeb-DF}} &       & \multicolumn{2}{c}{\textbf{DFD}} &       & \multicolumn{2}{c}{\textbf{DFDC}} \\
\cline{3-4}\cline{6-7}\cline{9-10}\cline{12-13}          &       & \textbf{AUC} & \textbf{EER$\downarrow$} &       & \textbf{AUC} & \textbf{EER$\downarrow$} &       & \textbf{AUC} & \textbf{EER$\downarrow$} &       & \textbf{AUC} & \textbf{EER$\downarrow$} \\
    \hline
    & Xception$^\dagger$ & 0.6059  & 0.6191  &       & 0.6005  & 0.4319 
&       & 0.6543 & 0.3933       &       & 0.5565   &0.4605   \\
& Add-Net$^\dagger$ &0.5421  &0.4621   &   &0.5783 &  0.4444     &       & 0.5716      &  0.4531     &       & 0.5160      &0.5477   \\ 
FF++ (LQ)  & F3-Net$^\dagger$&0.6049       & 0.4341      &       & 0.6795      &  	0.3676     &       &0.6950       & 0.3539       &       & 0.5787      &0.4423   \\  
& Multi-Att$^\dagger$& 0.6565  & 0.3965  &       & 0.6864 &	0.3708 &       & 0.7418   & 0.3272  &       & 0.6302 	& 0.4098 \\ 	
& Ours  & \textbf{0.6739 } & \textbf{0.3825 } &       & \textbf{0.6918 } & \textbf{0.3569 } &       & \textbf{0.7586 } & \textbf{0.3084 } &       & \textbf{0.6331}  & \textbf{0.4043} 
\\
    \hline
    \multirow{3}{*}{\tabincell{c}{\textbf{Training}\\\textbf{dataset}} } & \multirow{3}{*}{\textbf{Methods}} & \multicolumn{11}{c}{\textbf{Testing dataset}} \\
\cline{3-13}          &       & \multicolumn{2}{c}{\textbf{FF++(LQ)}} &       & \multicolumn{2}{c}{\textbf{Celeb-DF}} &       & \multicolumn{2}{c}{\textbf{DFD}} &       & \multicolumn{2}{c}{\textbf{DFDC}} \\
\cline{3-4}\cline{6-7}\cline{9-10}\cline{12-13}          &       & \textbf{AUC} & \textbf{EER$\downarrow$} &       & \textbf{AUC} & \textbf{EER$\downarrow$} &       & \textbf{AUC} & \textbf{EER$\downarrow$} &       & \textbf{AUC} & \textbf{EER$\downarrow$} \\
    \hline
    & Xception$^\dagger$ & 0.5920  &	0.4306
       &       &0.7791  &0.2944 &       & 0.8114 &0.2620 &       &0.5702       & 0.4510 \\
& Add-Net$^\dagger$&0.5388 & 0.4720 &       & 0.6212 & 0.4151 &       & 0.6877 & 0.3682 &       & 0.5337 & 0.4732  \\
WildDeepfake      & F3-Net$^\dagger$& 0.5595 & 0.4549 &       & 0.6088 & 0.4276 &       & 0.7827 & 0.2878 &       & 0.5240 & 0.4842 \\
& Multi-Att$^\dagger$& \textbf{0.6276} & \textbf{0.4097}  &       &0.7695   &	0.2811 
       &       &0.8416    & 0.2374   &       &0.5665   &0.4534 \\
& Ours  & 0.6160 & 0.4179 &       & \textbf{0.8294} & \textbf{0.2424} &       & \textbf{0.8680} & \textbf{0.1997} &       &\textbf{0.5894} 	&\textbf{0.4300} \\
    \hline
    \end{tabular}%
    \caption{Quantitative results on unseen datasets.}
    \label{tab:cross_testing}%
\end{table*}%

\noindent\textbf{Implementation detail.}
We implement the proposed framework via open-source PyTorch~\cite{pytorch}. We sample 50 frames per video at equal interval, then use the state-of-the-art face extractor DSFD~\cite{li2019dsfd} to detect faces and resize them to $320 \times 320$ (training of FF++) or $224 \times 224$ (training on WildDeepfake). The EffcientNet-B4~\cite{efficientnet} pre-trained on ImageNet was adopted as the backbone of our network, which is trained with Adam optimizer with the learning rate of $2 \times 10^{-4}$, the weight decay of $1 \times 10^{-5}$, and the batch size of 32. The stride of the sliding window is set to $2$ in all experiments. The results of comparisons are obtained from their paper,  and we specify our implementation by $^\dagger$ otherwise.

\noindent\textbf{Evaluation Metrics.} Following the convention~\cite{ffpp, f3net, df_mask}, we apply Accuracy score (Acc), Area Under the Receiver Operating Characteristic Curve (AUC) and Equal Error Rate (EER) as our evaluation metrics.

\noindent\textbf{Comparing Methods.}
We compare our method with several advanced methods: 
C-Conv\cite{bayar2016deep}, CP-CNN\cite{CP-CNN}, 
MesoNet \cite{mesonet}, Xception \cite{xception}, Face X-ray \cite{face_xray}, Two-branch RN \cite{two_branch_rn}, Add-Net \cite{wild_deepfake}, F$^3$-Net \cite{f3net}, FDFL \cite{CVPR2021FDFL} and Multi-Att \cite{cvpr2021multi}.

\subsection{Experimental Results}
\subsubsection{Intra-testing.}
Following~\cite{ffpp, f3net}, we compare our method against several advanced techniques on the FF++ dataset with different quality settings (\emph{i.e.}, HQ and LQ), and further evaluate the effectiveness of our approach on the WlidDeepfake dataset that is closer to the real-world scenario.
The quantitative results are shown in Tab.\ref{tab:ffpp}. It can be observed that our proposed method outperforms all the compared methods in both Acc and AUC metrics across all dataset settings.
These gains mainly benefit from the proposed PEL framework that can sufficiently excavate and magnify the fine-grained forgery cues.% hidden in the fake faces.

\begin{table*}[htbp]
  \centering
    \begin{tabular}{cl|cccccccc}
    \hline
    \multirow{2}{*}{\textbf{Datasets}} & \multirow{2}{*}{\textbf{Methods}} & \multicolumn{2}{c}{\textbf{+ GaussianNoise}} &       & \multicolumn{2}{c}{\textbf{+ SaltPepperNoise}} &       & \multicolumn{2}{c}{\textbf{+ GaussianBlur}} \\
\cline{3-4}\cline{6-7}\cline{9-10}          &       & \textbf{$\Delta$Acc} & \textbf{$\Delta$AUC} &       & \textbf{$\Delta$Acc} & \textbf{$\Delta$AUC} &       & \textbf{$\Delta$Acc} & \textbf{$\Delta$AUC} \\
    \hline
    & Xception \cite{xception}  & -2.65\% & -0.0397 &       & -32.44\% & -0.3330 &       & \textbf{-6.22\%} & -0.1994 \\
& AddNet$^\dagger$ \cite{wild_deepfake} & -41.51\% & -0.2862 &       & -11.28\% & -0.3445 &       & -11.28\% & -0.3445 \\
FF++ (LQ)    & F3Net$^\dagger$ \cite{f3net} & -9.86\% & -0.0838 &       & -31.08\% & -0.3891 &       & -11.08\% & -0.2077 \\
& Multi-att$^\dagger$ \cite{cvpr2021multi} & -1.79\% & -0.0058  &       & -49.30\% & -0.2494 &       & -12.23\% & -0.2475 \\
& Ours  & \textbf{-0.10\%} & \textbf{-0.0031} &       & \textbf{-9.39\%} & \textbf{-0.2079} &       & -7.41\% & \textbf{-0.1274} \\
    \hline
    & Xception \cite{xception}  & -0.98\% & -0.0082 &       & -27.80\% & -0.1373 &       & -12.71\% & \textbf{-0.0664} \\
& AddNet$^\dagger$ \cite{wild_deepfake} & -11.66\% & -0.3327 &       & -18.21\% & -0.3589 &       & -12.91\% & -0.1895 \\
WildDeepfake      & F3Net$^\dagger$ \cite{f3net} & -1.17\% & -0.0248 &       & -43.57\% & -0.3407 &       & -12.43\% & -0.1567 \\
& Multi-att$^\dagger$ \cite{cvpr2021multi} & -0.99\% &  -0.0139 &       & -29.47\% & -0.1813 &       & -14.86\% & -0.1829 \\
& Ours  & \textbf{-0.86\%} & \textbf{-0.0050} &       & \textbf{-4.25\%} & \textbf{-0.0483} &       & \textbf{-10.88\%} & -0.1216 \\
    \hline
    \end{tabular}%
    \caption{Robustness evaluation under various types of perturbations.}
  \label{tab:robust}%
\end{table*}%

\subsubsection{Cross-testing.} 
As new forgery methods are emerging all the time, the generalizability of detection models directly affects their application under real-world scenarios. 
To test the generalization ability, we perform cross-dataset evaluation, \textit{i.e.}, training on one dataset while testing on another different dataset. This is more challenging since the testing distribution is different from that of training.
We compare our method against four competing methods through cross-dataset evaluations, as shown in Tab.~\ref{tab:cross_testing}. Specifically, we first trained these models on FF++ (LQ) and then evaluated their AUC and EER metrics on WildDeepfake, Celeb-DF, DFD and DFDC, respectively. 
From Tab.~\ref{tab:cross_testing} we can observe that our method outperforms all the competitors significantly on all testing datasets.
We further trained our model on WildDeepfake and evaluated it on the other four datasets, obtaining similar results that our method substantially outperforms the compared methods in nearly all cases. 

\noindent\textbf{Robustness.} 
In the process of video acquisition and transmission, various noises such as blur or salt will be introduced to the digital data; hence the robustness towards perturbation of the detection model is essential to the real-world application.
Thus, it's important for a model to be robust to all kinds of perturbation.
A series of experiments are conducted to verify the robustness of our method.
Specifically, we apply Gaussian noise, salt and pepper noise, and Gaussian blur to the test data, and measure the decay of Acc and AUC (denote as $\Delta$Acc and $\Delta$AUC) to assess the robustness of the detection model, as shown in Tab.~\ref{tab:robust}. The results turn out that the performance of our method degrades the least in most cases, which mainly benefits from our novel noise enhancement mechanism.

\begin{table}[t]
  \centering
    \begin{tabular}{cccc|cc}
    \hline
    \textbf{RGB} & \textbf{Freq} & \textbf{Self} & \textbf{Mutual} & \textbf{Acc} & \textbf{AUC} \\
    \hline
    \checkmark      &       &       &       & 88.59\%	 & 0.9251  \\
    \hline
    & \checkmark      &       &       & 88.65\%	 & 0.9250  \\
    \hline
    \checkmark &     & \checkmark      & \checkmark      & 89.15\%	 & 0.9307  \\
    \hline
    & \checkmark     & \checkmark      &  \checkmark     & 89.11\%	 & 0.9261  \\
    \hline
    \checkmark     & \checkmark     &       &       & 89.51\%   & 0.9369 \\	

    \hline
    \checkmark     & \checkmark     & \checkmark     &       & 89.98\%  & 0.9390 \\	

    \hline
    \checkmark     & \checkmark     &       & \checkmark     & 89.98\%  & 0.9394 \\
    \hline
    \checkmark     & \checkmark     & \checkmark     & \checkmark     &  \textbf{90.52\%}    & \textbf{ 0.9428} \\
    \hline
    \end{tabular}%
    \caption{Abalation study on FaceForensics++ (LQ) dataset.}
  \label{tab:ablation}%
  \vspace{-0.15in}
\end{table}%

\subsection{Ablation study}

\noindent\textbf{Components.}
As shown in Tab.\ref{tab:ablation}, we develop several variants and conduct a series of experiments on the FF++ (LQ) dataset to explore the influence of different components in our proposed method. 
In the single-stream setting, using only the integral RGB data or the fine-grained frequency data as input leads to similar results, and our proposed two modules can enhance the performance a little (the mutual-enhancement module is replaced with a spatial attention). 
In the two-stream setting, combining the original two stream can slightly improve the performance, which verifies that the fine-grained frequency input is distinct and complementary to the RGB data. 
The performance can be improved by adding the proposed self-enhancement module or mutual-enhancement module, reaching the peak when using the overall PEL framework. This shows the effectiveness of each module:
the self-enhancement module excavates the essential information within each stream separately, and the mutual-enhancement module enhances the above information by integrating them.

% Table generated by Excel2LaTeX from sheet 'paper'
\begin{table}[t]
  \centering

    \begin{tabular}{l|cc}
    \hline
    \textbf{Denoising methods} & \textbf{Acc} & \textbf{AUC} \\
    \hline
    Median filter (ks = 3) & \textbf{90.52\%} & \textbf{0.9428} \\
    %\hline
    Mean filter (ks = 3) &  89.30\%   & 0.9346 \\	
    %\hline
    Non-local means (ks = 3) & 89.48\% & 0.9313\\ 
    %\hline
    \hline
    Median filter (ks = 5) & 89.25\% & 0.9356 \\
    %\hline
    Median filter (ks = 7) & 89.32\% & \multicolumn{1}{l}{0.9394} \\
    %\hline
    \hline
    Multiple median filter (ks = 3, 5) & 90.12\% & 0.9382 \\
    %\hline
    Multiple median filter (ks = 3, 5, 7) & 90.02\% & 0.9365 \\
    \hline
    \end{tabular}%
  \caption{Quantitative results of different denoising methods on FaceForensics++ (LQ) dataset. (ks: kernel size)}
  \label{tab:denoise}%
  \vspace{-0.15in}
\end{table}%

\noindent\textbf{Denoising methods.}
We seek a suitable denoising method by exploring filters with different types, kernel sizes and combinations. Tab.~\ref{tab:denoise} lists the result on FF++ (LQ) dataset.
For the filter type, we use the median filter, the mean filter and the non-local means~\cite{nlmeans}, in which the median filter beats the other two.
For the kernel size, we set it as $3,5,7$, respectively, and it turns out that the smallest kernel performs best.
We further combine multiple median filters to extract noise with different size regions, followed by $ 1 \times 1 $ group convolutional layer to reduce the number of channels. However, the single median filter seems to be better than multiple filters. Based on the above analysis,  we choose the single median filter with a kernel size of $3$ as the default denoising setting of our method under all experiments.

\begin{figure*}[t]
\centering
\includegraphics[scale=0.38]{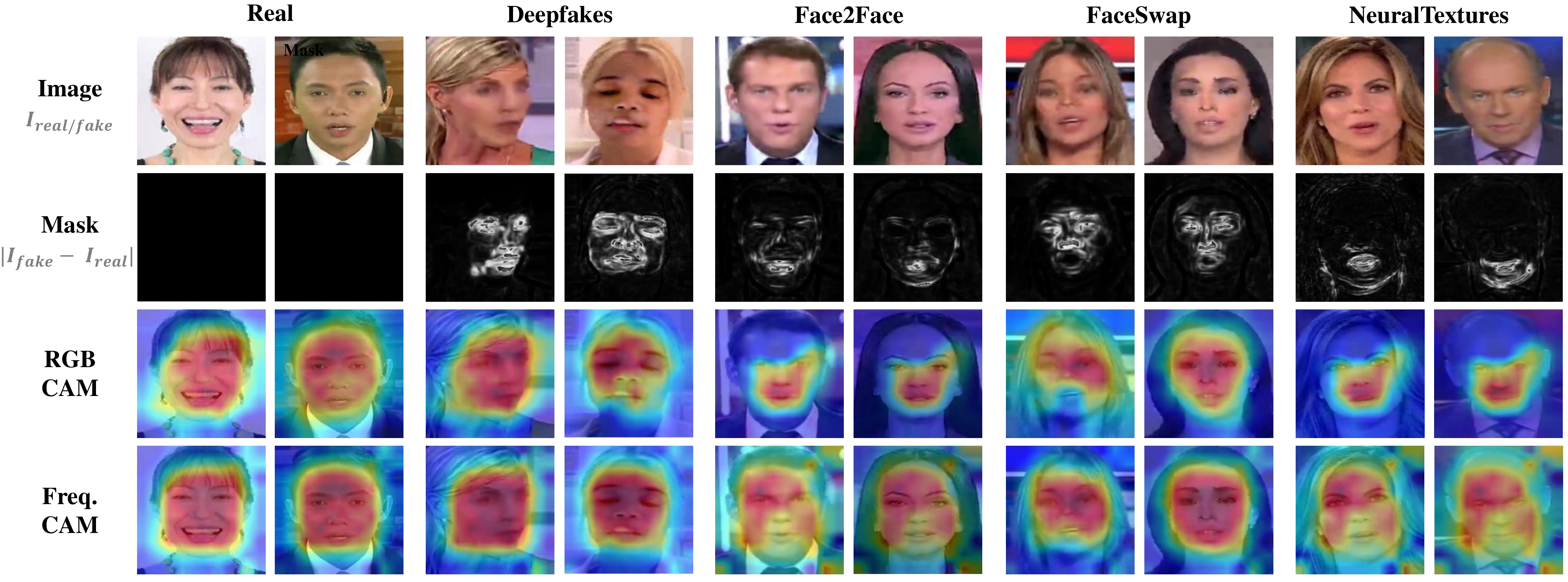}
\caption{The attention maps for different kinds of faces.}
\label{fig_cam}
\vspace{-0.1in}
\end{figure*}

\begin{figure}[t]
\centering
\includegraphics[scale=0.55]{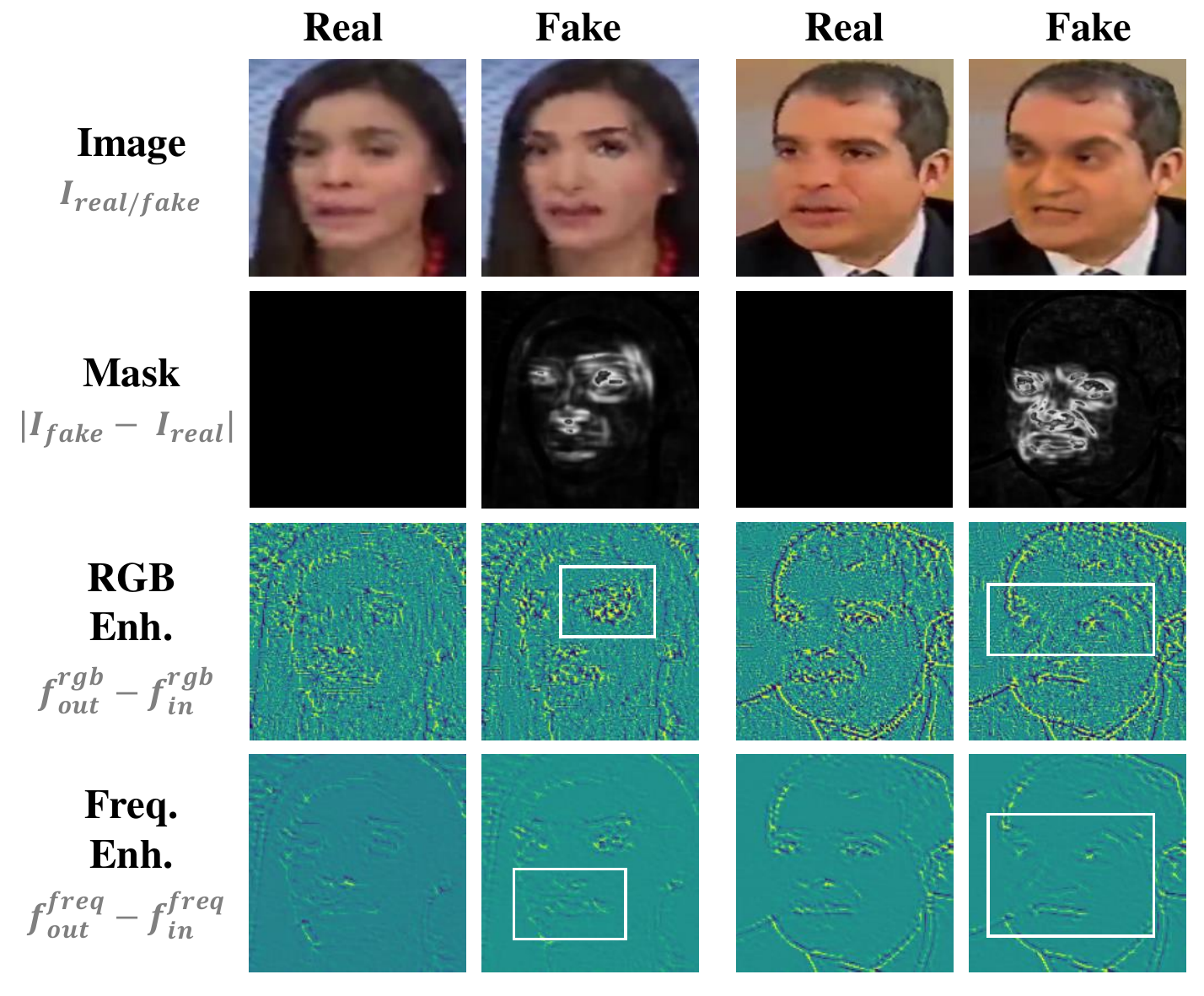}  %0.6
\caption{The self-enhancement of feature maps for the RGB stream and the frequency stream.}
\label{fig_vis_self}
\vspace{-0.15in}
\end{figure}

\begin{figure}[t]
\centering
\includegraphics[scale=0.55]{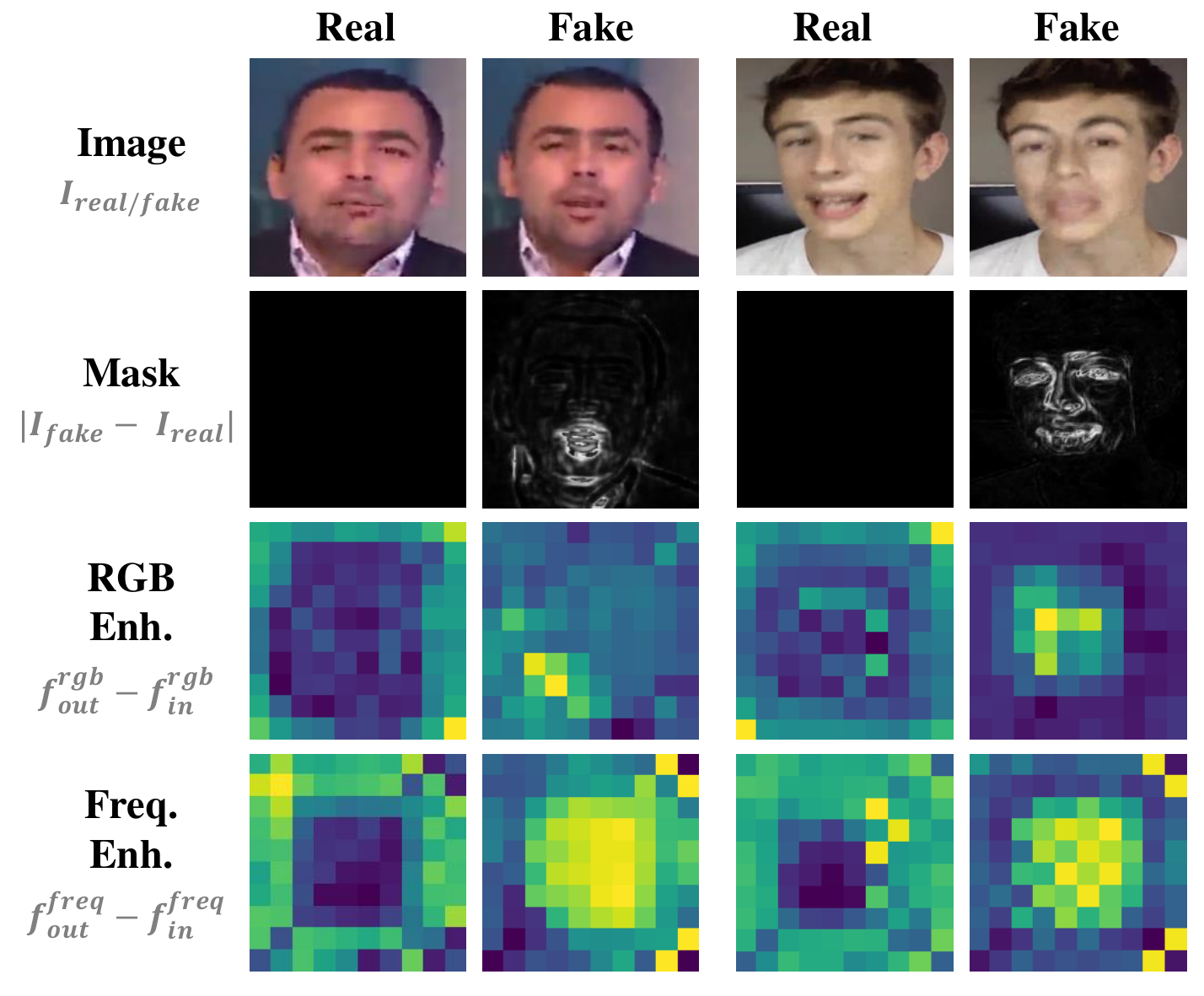}  
\caption{The mutual-enhancement of feature maps for the RGB stream and the frequency stream.}
\label{fig_vis_mutual}
\vspace{-0.15in}
\end{figure}

\subsection{Visualizations}

\noindent\textbf{Class Activation Mapping.}
To explore the region of interest for different face types, we visualized the attention maps of several samples by class activation mapping (CAM)~\cite{ZhouKLOT16}. As shown in Fig. \ref{fig_cam}, both RGB and frequency streams can locate the artifacts, but with different emphasis. The RGB stream process inputs in the original image space with relatively focused attention maps: for the forged images, the focus area mainly coincides with the mask of forged regions; while for real images, the attention is evenly focused. The frequency stream considers a larger area (\emph{i.e.} the entire face), which is often complementary to the focus area of the RGB stream, owing to the fine-grained frequency components' ability to reveal more subtle artifacts. In this manner, two streams can contribute to each other for a better forgery clues extraction.

\noindent\textbf{Self-enhancement.} 
Fig.~\ref{fig_vis_self} illustrates the enhanced region of feature maps (\emph{i.e.}, the residual between $f_{in}$ and $f_{out}$ in Sec.~\ref{sec_self}) via the first self-enhancement module, where mask denotes the forged region which is obtained from the pixel-level difference between a forged image and its corresponding original image.
From Fig.~\ref{fig_vis_self}, we can observe that
although the overall feature distributions of real and fake are similar, the latter manifests a higher enhancement in local manipulated regions, such as eyes or mouth.
Such a fine-grained self-enhancement mechanism enables our method to uncover subtle forgery cues.

\noindent\textbf{Mutual-enhancement.} 
Fig.\ref{fig_vis_mutual} illustrates the representation residual before and after the last mutual-enhancement module. With the cooperation of the Complementary RGB feature and frequency feature, the manipulated regions on both streams are enhanced simultaneously, increasing the discriminability between real and fake faces. 
Moreover, both streams suppress the facial region in real images while intensifying that in fake images, enhancing the distinctiveness of fine-grained clues. We also observe different enhanced regions between RGB and frequency streams, \emph{i.e.}, the former fit manipulated mask well, while the latter cover whole face with more subtle cues. The visualization illustrates that RGB and frequency information are independent but can benefit from each other.

\section{Conclusion}
In this paper, we propose a Progressive Enhancement Learning (PEL) framework to conduct the discriminative representation enhancement based on fine-grained frequency information.
Firstly, we decompose the image into fine-grained frequency components and fed it into a two-stream network together with the RGB input, which aids in the subsequent two enhancement modules.
The self-enhancement module captures the traces in different input spaces based on spatial noise enhancement and channel attention.
The mutual enhancement module achieves concurrent enhancement of both branches through feature communication in the shared spatial dimension. 
The PEL framework can fully exploit the fine-grained clues in high-quality forgery faces and achieves state-of-the-art performance on both seen and unseen datasets.
Visualizations of feature maps and class activation mappings reveal the inner mechanism and explain the effectiveness of our method.

\section*{Acknowledgement}
This work is supported by 
National Key Research and Development Program of China (No. 2019YFC1521104),
Shanghai Municipal Science and Technology Major Project (No. 2021SHZDZX0102), 
Shanghai Science and Technology Commission (No. 21511101200),
Zhejiang Lab (No. 2020NB0AB01),
National Natural Science Foundation of China (No. 61972157 and No. 72192821) and
Art major project of National Social Science Fund (No. I8ZD22).

% \begin{acks}
% \end{acks}

%\newpage
\bibliography{longstring,egbib}


%\appendix
%\input{section/appendix}

\end{document}